\documentclass{article} 
\usepackage{times}  
\usepackage{helvet}  
\usepackage{courier}  
\usepackage{url}  
\usepackage{graphicx}  

\usepackage{microtype}
\usepackage{subfigure}
\usepackage{booktabs} 
\usepackage{bbm}
\usepackage{bm}
\usepackage{amsmath}
\usepackage{amssymb}

\usepackage{algorithm}
\usepackage{algorithmic}

\usepackage{hyperref}
\usepackage{xspace}

\usepackage{color}

\newcommand{\mb}[1]{\mathbf{#1}}

\newcommand{\fullname}{Attentive Tensor Product Learning\xspace}
\newcommand{\name}{ATPL\xspace}
\newcommand{\eat}[1]{}

\newcommand{\ls}[1]  
   {\dimen0=\fontdimen6\the=#1\dimen0
    \advance\lineskip.5\fontdimen5\the\lineskip-\dimen0
    \lineskiplimit=.9\lineskip
    \baselineskip=\lineskip
    \advance\baselineskip\dimen0
    \normallineskip\lineskip
    \normallineskiplimit\lineskiplimit
    \normalbaselineskip\baselineskip
    \ignorespaces
   }

 \date{}

\begin{document}
\title{Attentive Tensor Product Learning}

\author{Qiuyuan Huang, Li Deng, Dapeng Wu, Chang Liu, Xiaodong He
\thanks{QH is with Microsoft Research AI, Redmond, WA; email: qihua@microsoft.com. LD is with Citadel; email: l.deng@ieee.org.  DW is with
University of Florida, Gainesville, FL 32611; email: dpwu@ufl.edu.
CL is with University of California, Berkeley; email:
liuchang@eecs.berkeley.edu. XH is with JD AI Research, Beijing,
China; email: xiaohe.ai@outlook.com.} }


\maketitle

\begin{abstract}
This paper proposes a new architecture --- \fullname (\name)
--- to represent grammatical structures in deep
learning models. \name exploits Tensor Product Representations (TPR), a structured
neural-symbolic model developed in cognitive science, to
integrate deep learning with explicit language structures and
rules. The key ideas of \name are: 1) unsupervised learning of
role-unbinding vectors of words via TPR-based deep neural network;
2) employing attention modules to compute TPR; and 3) integration
of TPR with typical deep learning architectures including Long
Short-Term Memory (LSTM) and Feedforward Neural Network (FFNN).
The novelty of our approach lies in its ability to extract the
grammatical structure of a sentence by using role-unbinding
vectors, which are obtained in an unsupervised manner. This \name
approach is applied to 1) image captioning, 2) part of speech
(POS) tagging, and 3) constituency parsing of a sentence.
Experimental results demonstrate the effectiveness of the proposed
approach.

\end{abstract}

\section{Introduction}

Deep learning (DL) is an important tool in many natural language
processing (NLP) applications. Since natural languages are rich in
grammatical structures, there is an increasing interest in
learning a vector representation to capture the grammatical
structures of the natural language descriptions using deep
learning models \cite{tai2015improved,kumar2016ask,kong2017dragnn}.

In this work, we propose a new architecture, called {\it \fullname
(\name)}, to address this representation problem by exploiting
Tensor Product Representations (TPR) \cite{smolensky1990tensor,smolensky2006harmonic}. TPR is a
structured neural-symbolic model developed in cognitive science
over 20 years ago. In the TPR theory, a sentence can be considered
as a sequences of {\it roles} (i.e., grammatical components) with
each filled with a {\it filler} (i.e., tokens). Given each role
associated with a {\it role vector} $r_t$ and each filler
associated with a {\it filler vector} $f_t$, the TPR of a sentence
can be computed as $S=\sum_t f_tr_t^\top$. Comparing with the
popular RNN-based representations of a sentence, a good property
of TPR is that decoding a token of a timestamp $t$ can be computed
directly by providing an {\it unbinding vector} $u_t$. That is,
$f_t=S\cdot u_t$. Under the TPR theory, encoding and decoding a
sentence is equivalent to learning the role vectors $r_t$ or
unbinding vectors $u_t$ at each position $t$.

We employ the TPR theory to develop  a novel attention-based
neural network architecture for learning the unbinding vectors
$u_t$ to serve the core at \name. That is, \name employs a form of
the recurrent neural network to produce $u_t$ one at a time. In
each time, the TPR of the partial prefix of the sentence up to
time $t-1$ is leveraged to compute the attention maps, which are
then used to compute the TPR $S_t$ as well as the unbinding vector
$u_t$ at time $t$. In doing so, our \name can not only be used to
generate a sequence of tokens, but also be used to generate a
sequence of {\it roles}, which can interpret the
syntactic/semantic structures of the sentence.

To demonstrate the effectiveness of our \name architecture, we
apply it to three important NLP tasks: 1) image
captioning; 2) POS tagging; and 3) constituency parsing of a
sentence. The first showcases our \name-based generator, while the
later two are used to demonstrate the power of role vectors in
interpreting sentences' syntactic structures. Our evaluation shows
that on both image captioning and POS tagging, our approach can
outperform previous state-of-the-art approaches. In particular, on
the constituency parsing task, when the structural segmentation is given as a
ground truth, our \name approach can beat the state-of-the-art by
$3.5$ points to $4.4$ points on the Penn TreeBank dataset. These results
demonstrate that our ATPL is more effective at capturing the
syntactic structures of natural language sentences.


The paper is organized as follows. Section~\ref{sec:RelatedWork}
discusses related work. In Section~\ref{sec:GenArch}, we present
the design of \name. Section~\ref{sec:ImageCaptioning} through
Section~\ref{sec:ConstituencyParsing} describe three applications of \name, i.e., image
captioner, POS tagger, and constituency parser, respectively.
Section~\ref{sec:Conclusion} concludes the paper.

\section{Related work}
\label{sec:RelatedWork}


Our proposed image captioning system follows a great deal of
recent caption-generation literature in exploiting end-to-end deep
learning with a CNN image-analysis front end producing a
distributed representation that is then used to drive a
natural-language generation process, typically using RNNs
\cite{mao2015deep,vinyals2015show,karpathy2015deep}. Our
grammatical interpretation of the structural roles of words in
sentences makes contact with other work that incorporates deep
learning into grammatically-structured networks
\cite{tai2015improved,andreas2015deep,yogatama2016learning,maillard2017jointly}.
Here, the network is not itself structured to match the
grammatical structure of sentences being processed; the structure
is fixed, but is designed to support the learning of distributed
representations that incorporate structure internal to the
representations themselves
--- filler/role structure.

The second task we consider is POS tagging. Methods for automatic
POS tagging include unigram tagging, bigram tagging, tagging using
Hidden Markov Models (which are generative sequence models),
maximum entropy Markov models (which are discriminative sequence
models), rule-based tagging, and tagging using bidirectional
maximum entropy Markov models \cite{jurafsky2017Speech}. The
celebrated Stanford POS tagger of \cite{Stanford_Parser_weblink}
uses a bidirectional version of the maximum entropy Markov model
called a cyclic dependency network in \cite{toutanova2003feature}.

Methods for automatic constituency parsing of a sentence, our
third task, include methods based on probabilistic context-free
grammars (CFGs) \cite{jurafsky2017Speech}, the shift-reduce method
\cite{zhu2013fast}, sequence-to-sequence LSTMs
\cite{vinyals2015grammar}.  Our constituency parser is similar to
the sequence-to-sequence LSTMs \cite{vinyals2015grammar} since
both use LSTM neural networks to design a constituency parser.
Different from \cite{vinyals2015grammar}, our constituency parser
uses TPR and unbinding role vectors to extract features that
contain grammatical information.


\begin{figure*}[t]
    \centering
    \includegraphics[width=0.7\linewidth]{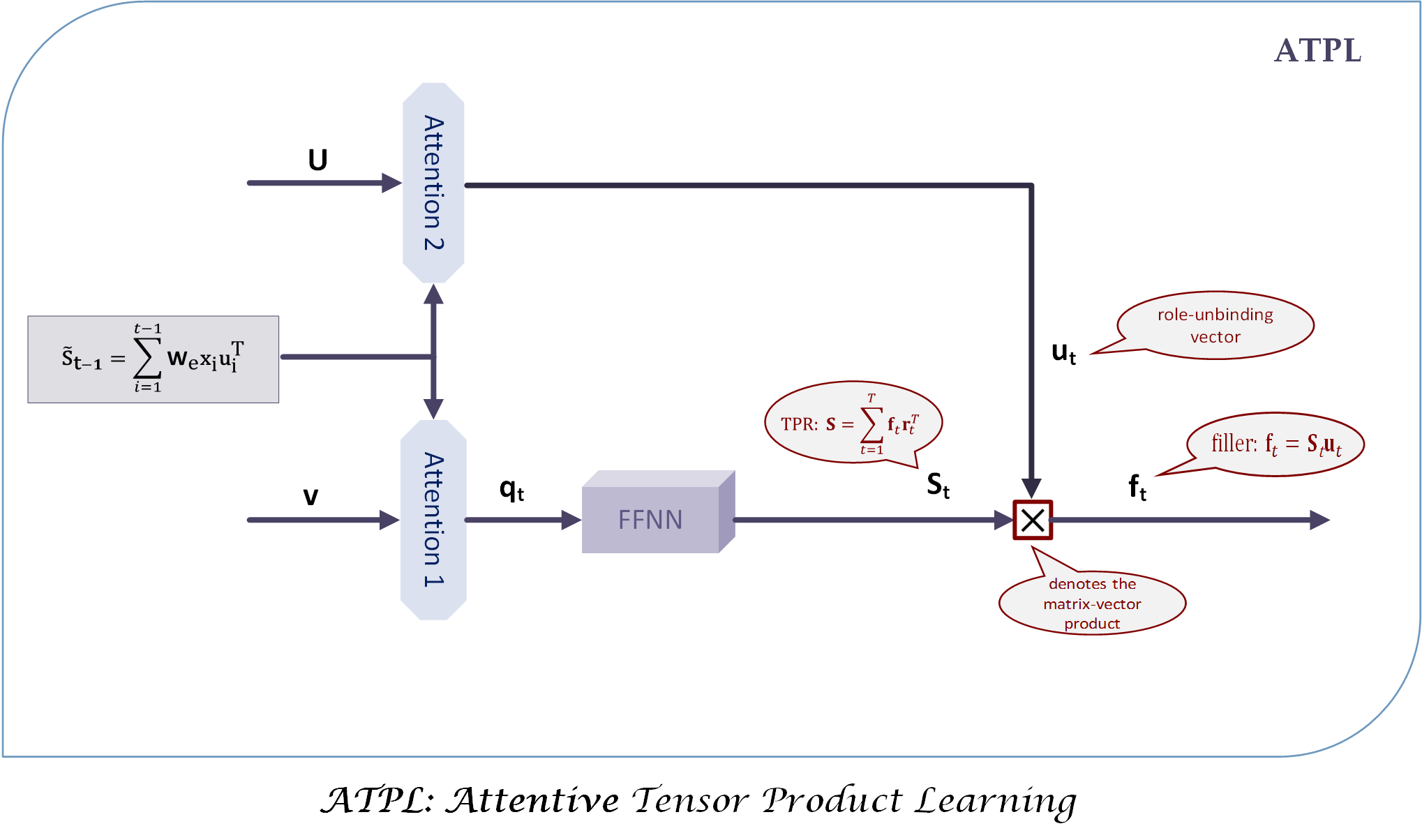}
    \caption{\name Architecture.}
    \label{fig:Architecture1}
\end{figure*}

\section{\fullname}
\label{sec:GenArch}

In this section, we present the \name architecture. We will first
briefly revisit the Tensor Product Representation (TPR) theory,
and  then introduce several building blocks. In the end, we
explain the \name architecture, which is illustrated in
Figure~\ref{fig:Architecture1}.

\subsection{Background: Tensor Product Representation}
\label{sec:naive-tpr}

The TPR theory allows computing a vector representation of a
sentence as the summation of its individual tokens while the order
of the tokens is within consideration. For a sentence of $T$ words, denoted by
$x_1,\cdots,x_T$, TPR theory considers the sentence as a sequence of
{\it grammatical role slots} with each slot filled with a concrete
token $x_t$. The role slot is thus referred to as a {\it role},
while the token $x_t$ is referred to as a {\it filler}.

The TPR of the sentence can thus be computed as {\it binding} each
role with a filler. Mathematically, each role is associated with a
{\it role vector} $r_t \in \mathbb{R}^{d}$, and a filler with a {\it filler vector}
$f_t \in \mathbb{R}^{d}$. Then the TPR of the sentence is
\begin{equation}
S=\sum_{t=1}^T f_t\cdot r_t^\top\label{eq:binding}
\end{equation}
where $S \in \mathbb{R}^{d\times d}$.
Each role is also associated with a dual {\it unbinding vector}
  $u_t$ so that $r_t^\top u_t=1$ and $r_t^\top u_{t'} = 0, t' \neq t$; then
\begin{equation}
f_t=Su_t\label{eq:unbinding}
\end{equation}
Intuitively, Eq.~\eqref{eq:unbinding} requires that $R^\top
U=\mathbf{I}$, where $R=[r_1;\cdots;r_T]$, $U=[u_1;\cdots;u_T]$, and $\mathbf{I}$ is an identity matrix. In a
simplified case, i.e., $r_t$ is orthogonal to each other and
$r_t^\top r_t=1$, we can easily derive $u_t=r_t$.

Eq. (\ref{eq:binding}) and (\ref{eq:unbinding}) provide means to
{\it binding} or {\it unbinding} a TPR. Through these mechanisms,
it is also easy to construct an encoder and a decoder to convert
between a sentence and its TPR. All we need to compute is the role
vector $r_t$ (or its dual unbinding vector $u_t$) at each timestep
$t$. One simple approach is to compute it as the hidden states of
a recurrent neural network (e.g., LSTM). However, this simple
strategy may not yield the best performance.

\subsection{Building blocks}

Before we start introducing \name, we first introduce several
building blocks repeatedly used in our construction.

\newcommand{\attn}[1]{\ensuremath{\mathrm{Attn}(#1)}}
An {\it attention module} over an input vector $v$ is defined as
\begin{equation}
\attn{v} = \sigma(Wv+b)
\end{equation}
where $\sigma$ is the sigmoid function, $W\in\mathbb{R}^{d_1\times
d_2}$, $b\in\mathbb{R}^{d_1}$, $d_2$ is the dimension of $v$, and
$d_1$ is the dimension of the output. Intuitively, $\attn{\cdot}$
will output a vector as the attention heatmap; and $d_1$ is equal
to the dimension that the heatmap will be attended to. $W$ and $b$
are two sets of parameters. Without specific notices, the sets of
parameters of different attention modules are disjoint to each
other.

\newcommand{\ffnn}[1]{\ensuremath{\mathrm{FFNN}(#1)}}
We refer to a {\it Feed-Forward Neural Network} (FFNN) module as a
single fully-connected layer:
\begin{equation}
\ffnn{v}=\mathbf{tanh}(Wv + b)
\end{equation}
where $W$ and $b$ are the parameter matrix and the parameter
vector with appropriate dimensions respectively, and
$\mathbf{tanh}$ is the hyperbolic tangent function.

\subsection{\name architecture}

In this paper, we mainly focus on an \name decoder architecture
that can decode a vector representation $\mathbf{v}$ into a
sequence $x_1,\cdots,x_T$. The architecture is illustrated in
Fig.~\ref{fig:Architecture1}.


We notice that, if we require the role vectors to be orthogonal to
each other, then to decode the filler $f_t$ only needs to unbind the TPR of undecoded words, $S_t$:
\begin{equation}
f_t = S_tu_t=\big(\sum_{i=t}^T (W_ex_i)r_i^\top\big) u_t = W_ex_t
\label{eq:decoder}
\end{equation}
where $x_t\in \mathbb{R}^V$ is a one-hot
encoding vector of dimension $V$ and $V$ is the size of the
vocabulary; $W_e \in \mathbb{R}^{d\times V}$ is a
word embedding matrix, the $i$-th column of which is the embedding
vector of the $i$-th word in the vocabulary; the embedding vectors are obtained by the
Stanford GLoVe algorithm with zero mean
\cite{Stanford_Glove_weblink}.

To compute $S_t$ and $u_t$, \name employs two attention modules  controlled by $\tilde{S}_{t-1}$, which is the TPR of the so-far
generated words $x_1,\cdots,x_{t-1}$:
\[\tilde{S}_{t-1} = \sum_{i=1}^{t-1} W_ex_i{r}_i^\top\]

On one hand, $S_t$ is computed as follows:
\begin{eqnarray}
S_t = \ffnn{\mathbf{q}_t}\\
\mathbf{q}_t=\mathbf{v}\odot\attn{h_{t-1}\oplus\mathrm{vec}(\tilde{S}_{t-1})}
\end{eqnarray}
where $\odot$ is the point-wise multiplication, $\oplus$
concatenates two vectors, and ${\mathrm vec}$ vectorizes a matrix.
In this construction, $h_{t-1}$ is the hidden state of an external
LSTM, which we will explain later.

The key idea here is that we employ an attention model to put
weights on each dimension of the image feature vector $\mathbf{v}$, so
that it can be used to compute $S_t$. Note it has been
demonstrated that that attention structures can be used to
effectively learn any function~\cite{vaswani2017attention}. Our
work adopts a similar idea to compute $S_t$ from $\mathbf{v}$ and
$\tilde{S}_{t-1}$.

On the other hand, similarly, $u_t$ is computed as follows:
\[u_t = \mathbf{U}\attn{h_{t-1}\oplus\mathrm{vec}(\tilde{S}_{t-1})}\]
where $\mathbf{U}$ is a constant normalized Hadamard matrix.

In doing so, \name can decode an image feature vector ${\mathbf v}$ by
recursively 1) computing $S_t$ and $u_t$ from $\tilde{S}_{t-1}$,
2) computing $f_t$ as $S_tu_t$, and 3) setting
${r}_t=u_t$ and updating $\tilde{S}_t$. This procedure
continues until the full sentence is generated.



\begin{figure*}[tbh]
    \centering
    \includegraphics[width=0.8\textwidth]{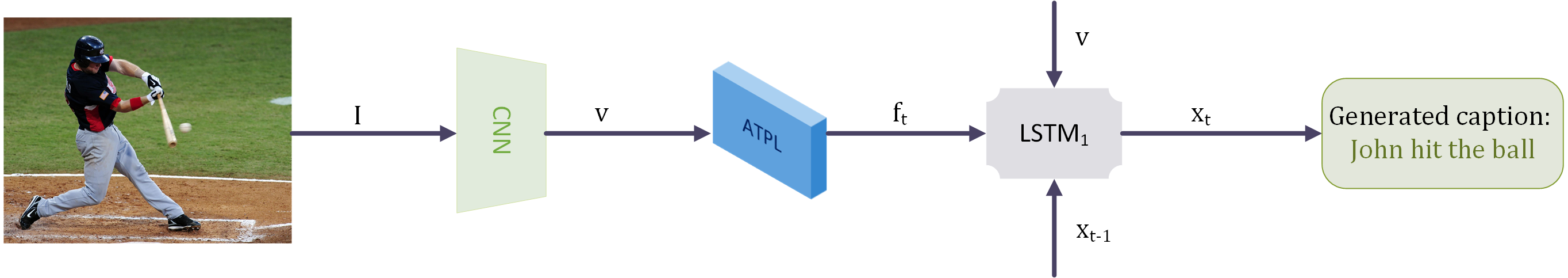}
    \caption{Architecture of image captioning.}
    \label{fig:Architecture3}
\end{figure*}

{\small
    \begin{table*}[t]
        \caption{Performance of the proposed \name model on the COCO dataset.}
        \label{table:BLEU}
        \centering
        \begin{tabular}{lllllllll}
            \hline
            Methods     & METEOR &BLEU-1 & BLEU-2 & BLEU-3 & BLEU-4 &   CIDEr\\
            \hline
            NIC \cite{vinyals2015show} & 0.237 &  0.666& 0.461& 0.329& 0.246&  0.855\\
            CNN-LSTM \cite{SCN_CVPR2017} &0.238 & 0.698 &0.525 &0.390 &0.292  & 0.889 \\
            SCN-LSTM \cite{SCN_CVPR2017} &0.257 & 0.728 &0.566 &0.433 &0.330  & 1.012 \\
            \name  & \textbf{0.258}& \textbf{0.733} & \textbf{0.572} &  \textbf{0.437} & \textbf{0.335}  &  \textbf{1.013}\\
            \hline
            \hline
        \end{tabular}
    \end{table*}
}

Next, we will present three applications of \name, i.e., image
captioner, POS tagger, and constituency parser in Section~\ref{sec:ImageCaptioning} through
Section~\ref{sec:ConstituencyParsing}, respectively.

\section{Image Captioning}
\label{sec:ImageCaptioning}

To showcase our \name architecture, we first study its application
in the image captioning task. Given an input image $\mathbf{I}$, a
standard encoder-decoder can be employed to convert the image into
an image feature vector $\mathbf{v}$, and then use the \name decoder to
convert it into a sentence. The overall architecture is dipected
in Fig.~\ref{fig:Architecture3}.

We evaluate our approach with several baselines on the COCO
dataset~\cite{COCO_weblink}. The COCO dataset contains 123,287
images, each of which is annotated with at least 5 captions. We
use the same pre-defined splits as
\cite{karpathy2015deep,SCN_CVPR2017}: 113,287 images for training,
5,000 images for validation, and 5,000 images for testing. We use
the same vocabulary as that employed in \cite{SCN_CVPR2017}, which
consists of 8,791 words.

For the CNN of Fig. \ref{fig:Architecture1}, we used ResNet-152
\cite{he2016deep}, pretrained on the ImageNet dataset. The
image feature vector ${\mathbf v}$ has 2048 dimensions.  The model is
implemented in TensorFlow \cite{tensorflow2015-whitepaper} with
the default settings for random initialization and optimization by
backpropagation. In our \name architecture, we choose $d=32$, and
the size of the LSTM hidden state to be $512$. The vocabulary size
$V=8,791$. \name uses tags as in \cite{SCN_CVPR2017}.

In comparison, we compare with \cite{vinyals2015show} and the
state-of-the-art CNN-LSTM and SCN-LSTM~\cite{SCN_CVPR2017}. The
main evaluation results on the MS COCO dataset are reported in
Table~\ref{table:BLEU}. The widely-used BLEU
\cite{papineni2002bleu}, METEOR  \cite{banerjee2005meteor}, and
CIDEr  \cite{vedantam2015cider} metrics are reported in our
quantitative evaluation of the performance of the proposed
scheme.

We can observe that, our \name architecture significantly
outperforms all other baseline approaches across all metrics being
considered. The results clearly attest to the effectiveness of the
\name architecture.
We attribute the performance gain of \name to the use of TPR in
replace of a pure LSTM decoder, which allows the decoder to learn
not only how to generate the {\it filler} sequence but also how to
generate the {\it role} sequence so that the decoder can better
understand the grammar of the considered language.
Indeed, by manually inspecting the generated captions from \name,
none of them has grammatical mistakes. We attribute this to the
fact that our TPR structure enables training to be more
effective and more efficient in learning the structure through the
role vectors.

Note that the focus of this paper is on developing a Tensor
Product Representation (TPR) inspired network to replace the core
layers in an LSTM; therefore, it is directly comparable to an LSTM
baseline. So in the experiments, we focus on comparison to a
strong CNN-LSTM baseline. We acknowledge that more recent papers
reported better performance on the task of image captioning. Performance improvements in these more recent
models are mainly due to using better image features such as those
obtained by Region-based Convolutional Neural Networks (R-CNN), or
using reinforcement learning (RL) to directly optimize metrics
such as CIDEr to provide a better context vector for caption
generation, or using an ensemble of multiple LSTMs, among others.
However, the LSTM is still playing a core role in these works and
we believe improvement over the core LSTM, in both performance and
interpretability, is still very valuable. Deploying these new features and
architectures (R-CNN, RL, and ensemble) with ATPL is
our future work.

\begin{figure*}[tbh]
    \centering
    \includegraphics[width=0.8\textwidth]{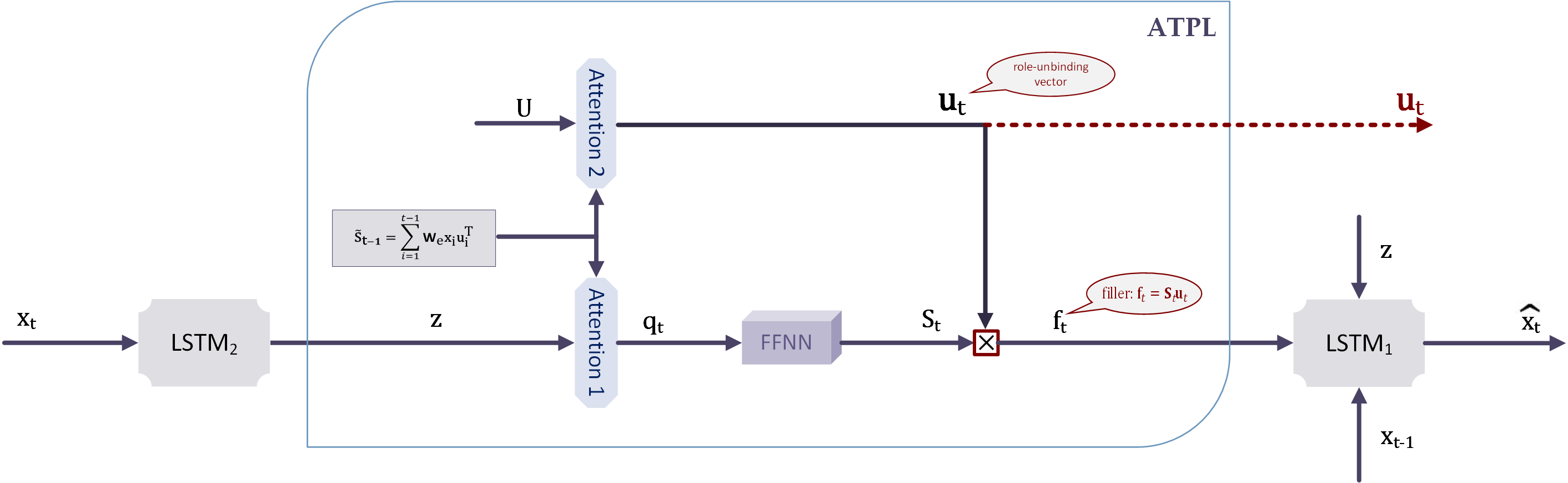}
    \caption{Architecture for acquisition of unbinding vectors of a sentence.}
    \label{fig:Architecture2}
\end{figure*}

\section{POS Tagging}
\label{sec:POS_Tagging}

In this section, we study the application of \name in the POS
tagging task. Intuitively, given a sentence $x_1,...,x_T$, POS tagging is to assign a POS tag
denoted as $z_t$, for each token $x_t$. In the following, we first
present our model using \name for POS tagging, and then evaluate
its performance.

\subsection{\name POS tagging architecture}

Based on TPR theory, the role vector (as well as its dual unbinding vector)  contains the POS tag
information of each word.
Hence, we first use \name to compute a
sequence of unbinding vectors $u_t$ which is of the same length as
the input sentence. Then we take $u_t$ and $x_t$ as input to a bidirectional LSTM model to produce a sequence of POS
tags.

Our training procedure consists of two steps. In the first step, we
employ an unsupervised learning approach to learn how to compute
$u_t$. Fig.~\ref{fig:Architecture2} shows a sequence-to-sequence structure, which uses an
LSTM as the encoder, and \name as the decoder; during the training phase of Fig.~\ref{fig:Architecture2}, the input is a sentence and the expected output is the same sentence as the input.
Then we use the trained system in Fig.~\ref{fig:Architecture2} to produce the
unbinding vectors $u_t$ for a given input sentence $x_1,...,x_T$.

\begin{figure}[t]
    \centering
    \includegraphics[width=0.3\textwidth]{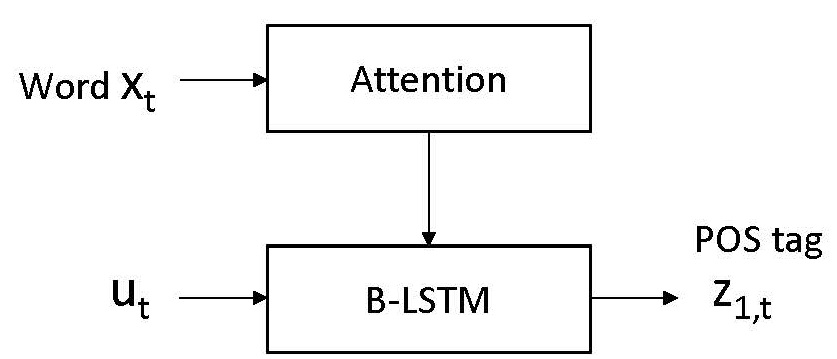}
    \caption{Structure of POS tagger.}
    \label{fig:POStagger}
\end{figure}

In the second step, we employ a bidirectional LSTM (B-LSTM) module
to convert the sequence of $u_t$ into a sequence of hidden states
$\mathbf{h}$. Then we compute a vector $z_{1,t}$ from each $(x_t,
\mathbf{h}_t)$ pair, which is the POS tag at position $t$. This procedure is illustrated in
Figure~\ref{fig:POStagger}.

The first step follows \name and is straightforward. Below, we
focus on explaining the second step. In particular, given the
input sequence $u_t$, we can compute the hidden states as
\begin{eqnarray}
\overrightarrow{{\mathbf h}}_{t},\overleftarrow{{\mathbf
        h}}_{t} = BLSTM(u_{t},\overrightarrow{{\mathbf
        h}}_{t-1},\overleftarrow{{\mathbf h}}_{t+1}) \label{eqn:BLSTM}
\end{eqnarray}

Then, the POS tag embedding is computed as

\begin{eqnarray}
{\mathbf z}_{1,t}= \mathbf{softmax}\big(\overrightarrow{{\mathbf
        W}}(x_t) \overrightarrow{{\mathbf h}}_{t} +
\overleftarrow{{\mathbf W}}(x_t) \overleftarrow{{\mathbf
h}}_{t}\big) \label{eq:TPR-LSTM3_5}
\end{eqnarray}

Here $\overrightarrow{{\mathbf W}}(x_t)$ is computed as follows

\begin{eqnarray}
\overrightarrow{{\mathbf W}}({\mathbf x})=
\overrightarrow{{\mathbf W}}_{a} \cdot
\textrm{diag}(\overrightarrow{{\mathbf W}}_{b}\cdot
    x_t) \cdot \overrightarrow{{\mathbf W}}_{c} \label{eqn:Wright}
\end{eqnarray}

where $\textrm{diag}(\cdot)$ constructs a diagonal matrix from the
input vector; $\overrightarrow{{\mathbf W}}_{a},
\overrightarrow{{\mathbf
        W}}_{b}, \overrightarrow{{\mathbf W}}_{c}$ are matrices of
appropriate dimensions. $\overleftarrow{{\mathbf
        W}}_{3,h}({\mathbf x}_t)$ is defined in the same manner as
$\overrightarrow{{\mathbf W}}_{3,h}({\mathbf x}_t)$, though a
different set of parameters is used.

Note that ${\mathbf z}_{1,t}$ is of dimension $P$, which is the total number of POS
tags.
Clearly, this model can be trained end-to-end by
minimizing a cross-entropy loss.

\subsection{Evaluation}

To evaluate the effectiveness of our model, we test it using the
Penn TreeBank dataset~\cite{Penn_treebank_weblink}. In particular,
we first train the sequence-to-sequence in
Fig.~\ref{fig:Architecture2} using the sentences of Wall Street
Journal (WSJ) Section 0 through Section 21 and Section 24 in Penn
TreeBank data set \cite{Penn_treebank_weblink}. Afterwards, we use
the same dataset to train the B-LSTM module in
Figure~\ref{fig:POStagger}.

\begin{table}[t]
    \caption{Performance of POS Tagger.  } \label{table:POStagger}
    \vskip 0.15in
    \begin{center}
        \begin{small}
            \begin{sc}
                \begin{tabular}{ccc|cc}
                    \toprule
                    & \multicolumn{2}{c|}{\cite{Stanford_Parser_weblink}} & \multicolumn{2}{c}{Our POS tagger} \\ \midrule
                    & WSJ 22 & WSJ 23 & WSJ 22 & WSJ 23 \\
                    Accuracy & 0.972 & 0.973   &\bf 0.973&\bf  0.974\\ \bottomrule
                \end{tabular}
            \end{sc}
        \end{small}
    \end{center}
\end{table}

Once the model gets trained, we test it on WSJ Section 22 and 23
respectively. We compare the accuracy of our approach against the
state-of-the-art Stanford parser~\cite{Stanford_Parser_weblink}.
The results are presented in Table~\ref{table:POStagger}. From the
table, we can observe that our approach outperforms the baseline.
This confirms our hypothesis that the unsupervisely trained
unbinding vector $u_t$ indeed captures grammatical information, so
as to be used to effectively predict grammar structures such as
POS tags.

{\scriptsize

}

\begin{figure}[tbh]
    \centering
    \includegraphics[width=0.4\textwidth]{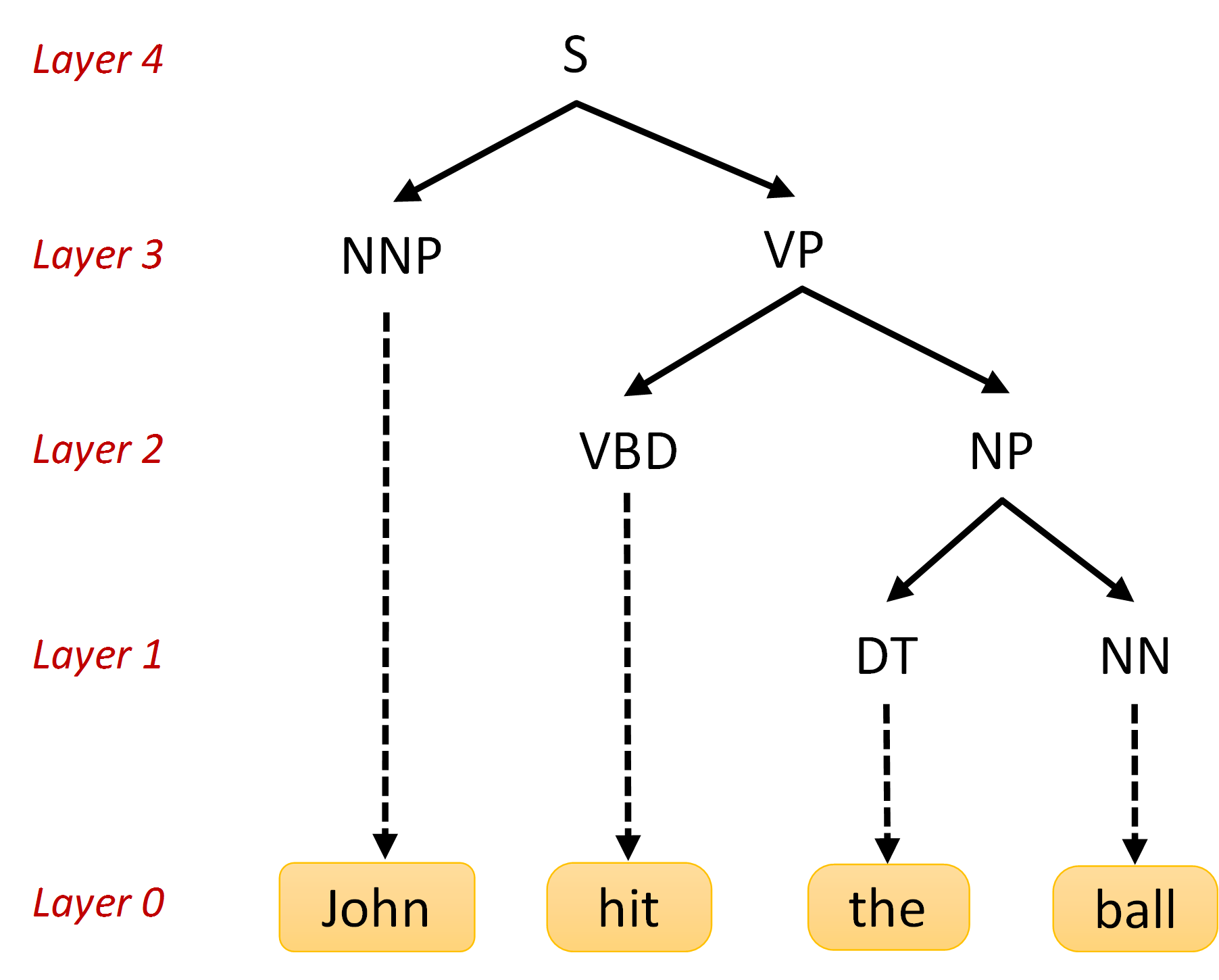}
    \caption{The parse tree of a sentence and its layers.}
    \label{fig:layers}
\end{figure}

\section{Constituency Parsing}
\label{sec:ConstituencyParsing}

In this section, we briefly review the constituency parsing task,
and then present our approach, which contains three component:
segmenter, classifier, and creator of a parse tree. In the end, we
compare our approach against the state-of-the-art approach
in~\cite{vinyals2015grammar}.


\subsection{A brief review of constituency parsing}

Constituency parsing converts a natural language into its parsing
tree. Fig.~\ref{fig:layers} provides an example of the parsing
tree on top of its corresponding sentence. From the tree, we can
label each node into layers, with the first layer (Layer 0)
consisting of all tokens from the original sentence. Layer $k$
contains all internal nodes whose depth with respect to the
closest leaf that it can reach is $k$.

In particular, at Layer 1 are all POS tags associated with each
token. In higher layers, each node corresponds to a {\it
substring}, a consecutive subsequence, of the sentence. Each node
corresponds to a grammar structure, such as a single word, a
phrase, or a clause, and is associated with a category. For
example, in Penn TreeBank, there are over 70 types of categories,
including (1) clause-level tags such as S (simple declarative
clause), (2) phrase-level tags such as NP (noun phrase), VP (verb
phrase), (3) word-level tags such as NNP (Proper noun, singular),
VBD (Verb, past tense), DT  (Determiner), NN  (Noun, singular or
mass), (4) punctuation marks, and (5) special symbols such as \$.

The task of constituency parsing recovers both the tree-structure
and the category associated with each node. In our approach to
employ \name to construct the parsing tree, we use an encoding $z$
to encode the tree-structure. Our approach first generates this
encoding from the raw sentence, layer-by-layer, and then predict a
category to each internal node. In the end, an algorithm is used
to convert the encoding $z$ with the categories into the full
parsing tree. In the following, we present the three sub-routines.

\subsection{Segmenting a sentence into a tree-encoding}
\label{subsec:segmenter}

We first introduce the concept of the encoding $z$. For each layer
$k$, we assign a value $\mb{z}_{k, t}$ to each location $t$ of the
input sentence. In the first layer, $\mb{z}_{1, t}$ simply encodes
the POS tag of input token $x_i$. In a higher level, $\mb{z}_{k,
t}$ is either $0$ or $1$. Thus the sequence $\mb{z}_{k, t}$ forms
a sequence with alternating sub-sequences of consecutive 0s and
consecutive 1s. Each of the longest consecutive 0s or consecutive
1s indicate one internal node at layer $k$, and the consecutive
positions form the substring of the node. For example, the second
layer of Fig.~\ref{fig:layers} is encoded as $\{0,1,0,0\}$, and the
third layer is encoded as $\{0, 1, 1, 1\}$.

The first component of our \name-based parser predicts $\mb{z}_{k,
t}$ layer-by-layer. Note that the first layer is simply the POS
tags, so we will not repeat it. In the following, we first explain
how to construct the second layer's encoding $\mb{z}_{2, t}$, and
then we show how it can be expanded to construct higher layer's
encoding $\mb{z}_{k, t}$ for $k\geq 3$.

\begin{figure}[t]
    \centering
    \includegraphics[width=0.4\textwidth]{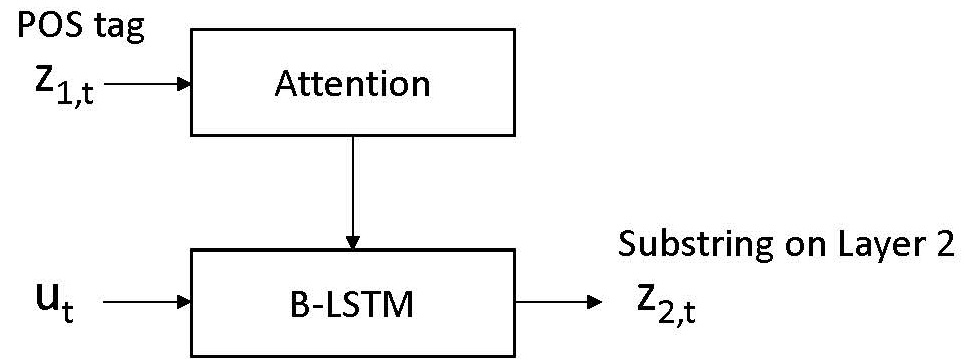}
    \caption{Structure of the segmenter on Layer 2.}
    \label{fig:substringLayer2}
\end{figure}

\paragraph{Constructing the second layer $\mb{z}_{2, t}$.} We can view $\mb{z}_{2, t}$ as a special tag over the POS tag sequence, and thus the same approach to compute the POS tag can be adapted here to compute $\mb{z}_{2, t}$. This model is illustrated in Fig.~\ref{fig:substringLayer2}.

In particular, we can compute the hidden state from the unbinding
vectors from the raw sentence as before:

\begin{eqnarray}
\overrightarrow{{\mathbf h}}_{2,t},\overleftarrow{{\mathbf
        h}}_{2,t} = BLSTM(u_{t},\overrightarrow{{\mathbf
        h}}_{2,t-1},\overleftarrow{{\mathbf h}}_{2,t+1})
\label{eqn:BLSTM2}
\end{eqnarray}

and the output of the attention-based B-LSTM is given as below
\begin{eqnarray}
{\mathbf z}_{2,t}= \sigma_s(\overrightarrow{{\mathbf
        W}}_{2}({\mathbf z}_{1,t}) \overrightarrow{{\mathbf h}}_{2,t} +
\overleftarrow{{\mathbf W}}_{2}({\mathbf z}_{1,t})
\overleftarrow{{\mathbf h}}_{2,t}) \label{eq:TPR-LSTM4_5}
\end{eqnarray}
where $\overrightarrow{{\mathbf W}}_{2,h}({\mathbf z}_{1,t})$ and
$\overleftarrow{{\mathbf W}}_{2,h}({\mathbf z}_{1,t})$ are defined
in the same manner as in \eqref{eqn:Wright}.

\begin{figure}[tbh]
    \centering
    \includegraphics[width=0.5\textwidth]{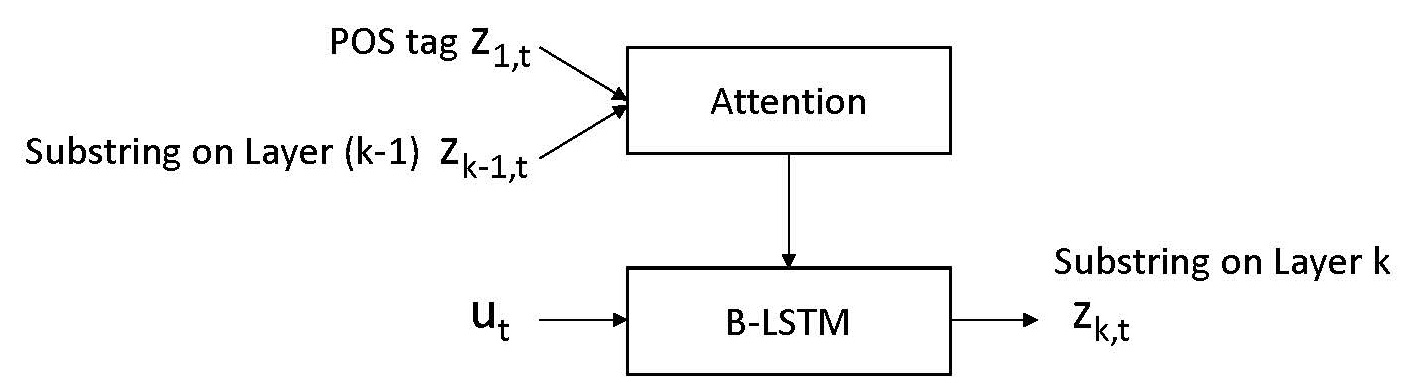}
    \caption{Structure of the segmenter on Layer $k\geq 3$.}
    \label{fig:substringLayer_k}
\end{figure}

\begin{figure}[tbh]
    \centering
    \includegraphics[width=0.4\textwidth]{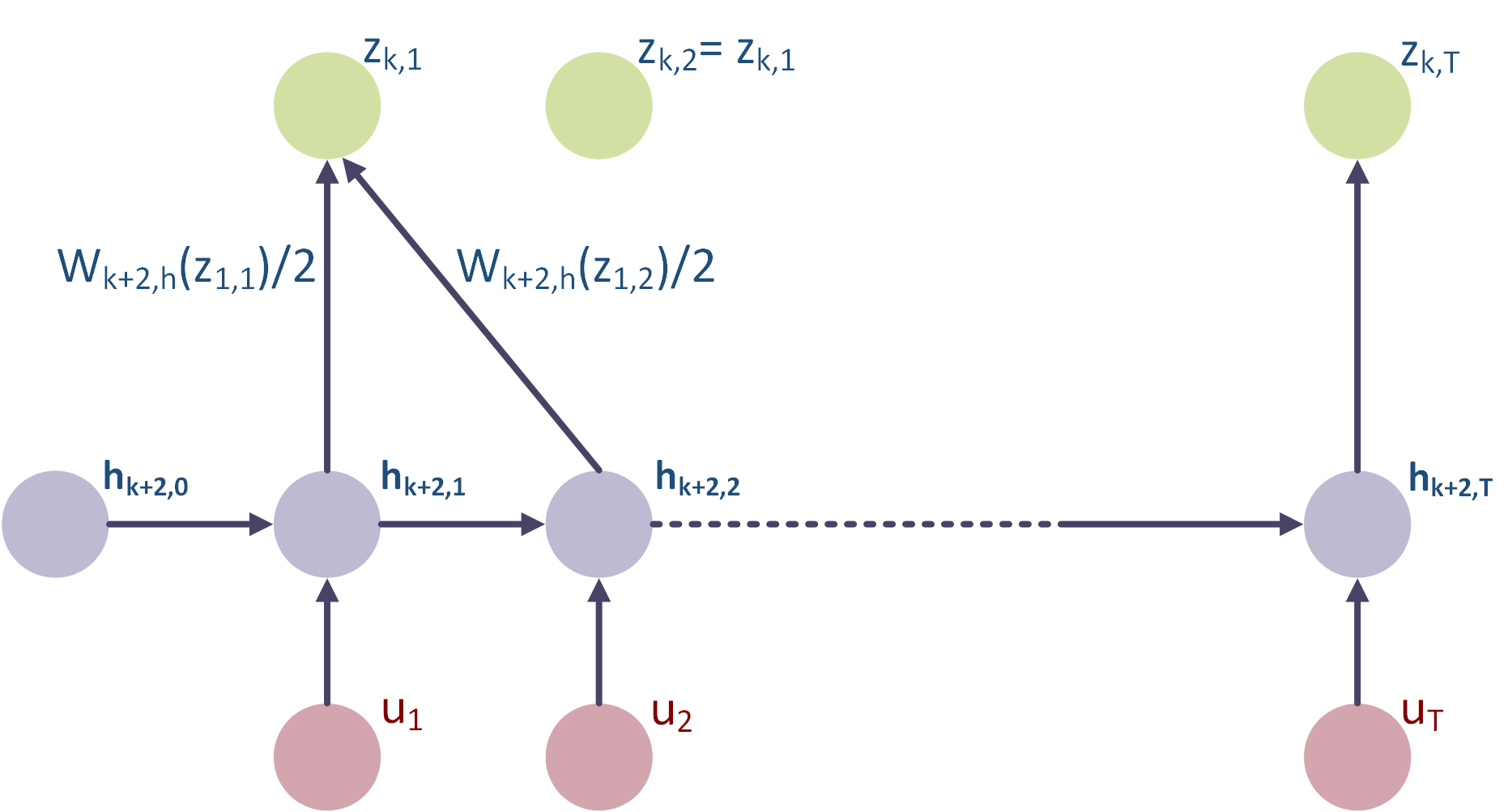}
    \caption{Segmenting Layer $k\geq 3$.}
    \label{fig:substringLayer_k_RNN}
\end{figure}

{\small
    \begin{table*}[t]
        \caption{Performance of Constituency Parser.  }
        \label{table:parseTree} \vskip 0.15in
        \begin{center}
            \begin{small}
                \begin{sc}
                    \begin{tabular}{ccc|cc|cc}
                        \toprule
                        & \multicolumn{2}{c|}{\cite{vinyals2015grammar}} & \multicolumn{2}{c|}{Our parser} & \multicolumn{2}{c}{Our parser with ground-truth ${\mathbf z}_{k,t}$ ($k\geq 2$)} \\ \midrule
                        & WSJ 22 & WSJ 23 & WSJ 22 & WSJ 23& WSJ 22 & WSJ 23\\
                        Precision & N/A  & N/A   & 0.898 &0.910  &0.952&0.952\\
                        Recall &N/A & N/A &0.901 &0.907 &0.973&0.978\\
                        F-1 measure & 0.928 & 0.921   &0.900& 0.908 & 0.963
                        & 0.965
                        \\ \bottomrule
                    \end{tabular}
                \end{sc}
            \end{small}
        \end{center}
    \end{table*}
}

\paragraph{Constructing higher layer's encoding $\mb{z}_{k, t}$ $(k\geq 3)$.}
Now we move to higher levels. For a layer $k\geq 3$, to predict
$\mb{z}_{k, t}$, our model takes both the POS tag input
$\mb{z}_{1, t}$ and the $(k-1)$-th layer's encoding $\mb{z}_{k-1,
t}$. The high-level architecture is illustrated in
Fig.~\ref{fig:substringLayer_k}.

Let us denote
\[\mb{z}_{k, t} = \mathbf{softmax}(J_{k, t})\]
the key difference is how to compute $J_{k, t}$. Intuitively,
$J_{k, t}$ is an embedding vector corresponding to the node, whose
substring contains token $x_t$. Assume word ${x}_t$ is in the
$m$-th substring of Layer $k-1$, which is denoted by $s_{k-1,m}$.
 Then, the embedding $J_{k, t}$ can be computed as
follows:
\begin{equation}
J_{k, t}=\sum_{i\in s_{k-1, m}} \frac{\overrightarrow{{\mathbf
            W}}_k({\mathbf z}_{1,i}) \overrightarrow{{\mathbf
            h}}_{k,i} + \overleftarrow{{\mathbf W}}_{k}({\mathbf
        z}_{1,i}) \overleftarrow{{\mathbf h}}_{k,i} }{|s_{k-1, m}|}
    \label{eq:bigJ}
\end{equation}
Here, $\overrightarrow{{\mathbf h}}_{k,i}$ and
$\overleftarrow{{\mathbf h}}_{k,i}$ are the hidden states of BLSTM
running over the unbinding vectors as before, and
$\overrightarrow{{\mathbf W}}_k(\cdot)$ and
$\overleftarrow{{\mathbf W}}_k(\cdot)$ are defined in a similar
fashion as (\ref{eqn:Wright}). We use $|\cdot|$ to indicate the
cardinality of a set.

The most interesting part is that $J_{k, t}$ aggregates all
embeddings computed from the substring of the previous layer
$s_{k-1, m}$. Note that the set $s_{k-1, m}$ of indexes can be
computed easily from $\mb{z}_{k-1, t}$. Note that many different
aggregation functions can be used. In (\ref{eq:bigJ}), we choose
to use the average function. The process of this calculuation is
illustrated in Fig.~\ref{fig:substringLayer_k_RNN}.

\begin{figure}[tbh]
    \centering
    \includegraphics[width=0.5\textwidth]{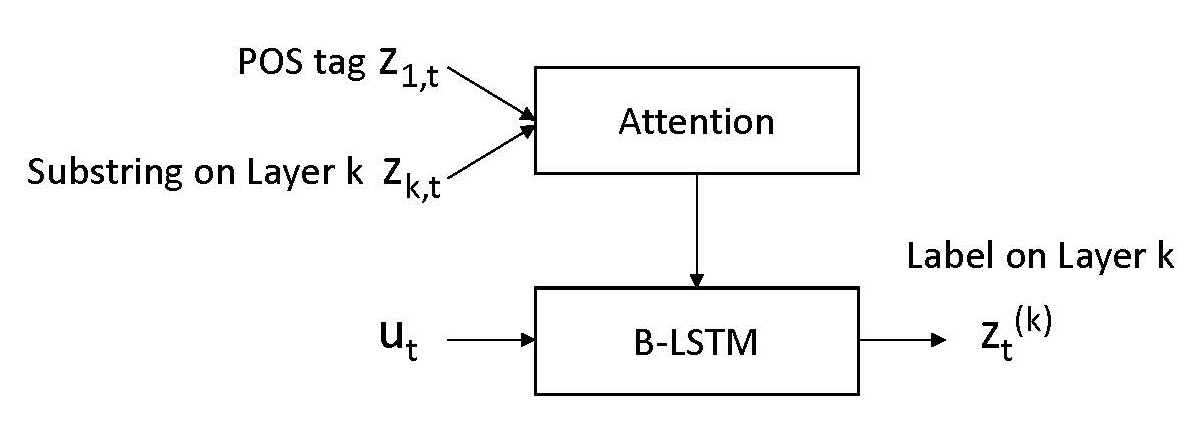}
    \caption{Structure of the classifier on Layer $k$.}
    \label{fig:classificationLayer_k}
\end{figure}


\subsection{Classification of Substrings}
\label{subsec:ClassificationSubstrings}

Once the tree structure is computed, we attach a category to each
internal node. We employ a similar approach as predicting
$\mb{z}_{k, t}$ for $k\geq 3$ to predict this category ${\mathbf
z}^{(k)}_t$. Note that, in this time, the encoding $\mb{z}_{k, t}$
of the internal node is already computed. Thus, instead of using
the encoding $\mb{z}_{k-1, t}$ from the previous layer, we use the
encoding of the current layer $\mb{z}_{k, t}$ to predict ${\mathbf
z}^{(k)}_t$ directly. This procedure is illustrated in
Fig.~\ref{fig:classificationLayer_k}.

Similar to (\ref{eq:bigJ}), we have ${\mathbf z}^{(k)}_t =
\mathbf{softmax}(E_{k, t})$, where $E_{k, t}$ is computed by
($\forall t \in \{t: {\mathbf x}_t \in s_{k,m}\}$)
\begin{equation}
E_{k, t}=\sum_{i\in s_{k, m}} \frac{\overrightarrow{{\mathbf
            W}}_k({\mathbf z}_{1,i}) \overrightarrow{{\mathbf
            h}}_{k,i} + \overleftarrow{{\mathbf W}}_{k}({\mathbf
        z}_{1,i}) \overleftarrow{{\mathbf h}}_{k,i} }{|s_{k, m}|}
\label{eq:bigE}
\end{equation}
Here, we slightly overload the variable names. We emphasize that
the parameters $\overrightarrow{{\mathbf W}}$ and
$\overleftarrow{{\mathbf W}}$  and the hidden states
$\overrightarrow{{\mathbf h}}_{k,i}$ and $\overleftarrow{{\mathbf
h}}_{k,i}$ are both independent to the ones used in
(\ref{eq:bigE}).

Note that the main different between (\ref{eq:bigE}) and
(\ref{eq:bigJ}) is that, the aggregation is operated over the set
$s_{k, t}$, i.e., the substring at layer $k$, rather than $s_{k-1,
t}$, i.e., the substring at layer $k-1$. Also, $E_{k, t}$'s
dimension is the same as the total number of categories, while
$J_{k, t}$'s dimension is 2.

\subsection{Creation of a Parse Tree}
\label{subsec:creator}

Once both $\mb{z}_{k, t}$ and ${\mathbf z}^{(k)}_t$ are
constructed, we can create the parse tree out of them using a
linear-time sub-routine. We rely on Algorithm~\ref{alg:parseTree}
to this end. For the example in Fig.~\ref{fig:layers}, the output
is (S(NNP John)(VP(VBD hit)(NP(DT the)(NN ball)))).

{\scriptsize
\begin{algorithm}[htb]
   \caption{Creation of a constituency parse tree}
   \label{alg:parseTree}
\begin{algorithmic}
   \STATE {\bf Input:} ${\mathbf x}_{t},{\mathbf
z}^{(k)}_t,{\mathbf z}_{k,t}$ ($t=1,\cdots,T$; $k=1,\cdots,h_p$)
 \STATE i=0
   \FOR{$j=1$ {\bfseries to} $h_p$}
    \FOR{$t=1$ {\bfseries to} $T$}
      \IF{$t=1$}
        \IF{$j==1$}
            \STATE output ``('' and ${\mathbf z}^{(h_p)}_1$
            \STATE push ${\mathbf z}^{(h_p)}_1$ into the stack
            \IF{${\mathbf z}^{(h_p)}_1=={\mathbf z}^{(1)}_1$}
                \STATE output   ${\mathbf x}_1$ and ``)''
                \STATE pop ${\mathbf z}^{(h_p)}_1$ out of the stack
            \ENDIF
        \ELSE
            \IF{${\mathbf z}^{(h_p-j+1)}_1\neq {\mathbf z}^{(h_p-j+2)}_1$}
                \STATE output ``('' and ${\mathbf z}^{(h_p-j+1)}_1$
                \STATE push ${\mathbf z}^{(h_p-j+1)}_1$ into the stack
                \IF{${\mathbf z}^{(h_p-j+1)}_1=={\mathbf z}^{(1)}_1$}
                    \STATE output   ${\mathbf x}_1$ and ``)''
                    \STATE pop ${\mathbf z}^{(h_p-j+1)}_1$ out of the stack
                \ENDIF
            \ENDIF
        \ENDIF
      \ELSE
        \IF{{\tiny ${\mathbf z}^{(h_p-j+1)}_t\neq {\mathbf z}^{(h_p-j+2)}_t$     \& ${\mathbf z}^{(h_p-j+1)}_t\neq {\mathbf z}^{(h_p-j+1)}_{t-1}$}}
            \STATE output ``('' and ${\mathbf z}^{(h_p-j+1)}_t$
            \STATE push ${\mathbf z}^{(h_p-j+1)}_t$ into the stack
            \IF{${\mathbf z}^{(h_p-j+1)}_t=={\mathbf z}^{(1)}_t$}
                \STATE output   ${\mathbf x}_t$ and ``)''
                \STATE pop ${\mathbf z}^{(h_p-j+1)}_t$ out of the stack
                \IF{$t==T$ or ${\mathbf z}^{(h_p-j+2)}_t\neq {\mathbf z}^{(h_p-j+2)}_{t+1}$}
                    \WHILE{the stack is not empty}
                        \STATE pop an element out of the stack
                        \IF{the substring of the element ends at
                        $t$} 
                            \STATE output    ``)''
                        \ELSE
                            \STATE push the element back into the stack
                        \ENDIF
                    \ENDWHILE
                \ENDIF
            \ENDIF
        \ENDIF
   \ENDIF
    \ENDFOR
   \ENDFOR
\end{algorithmic}
\end{algorithm}
}

\subsection{Evaluation}
\label{subsec:EvaluationParser}

We now evaluate our constituency parsing approach against the
state-of-the-art approach~\cite{vinyals2015grammar} using WSJ data
set in Penn TreeBank. Similar to our setup for POS tag, we
training our model using WSJ Section 0 through Section 21 and Section
24, and evaluate it on Section 22 and 23.

Table~\ref{table:parseTree} shows the performance for
both~\cite{vinyals2015grammar} and our proposed approach. In
addition, we also evaluate our approach assuming the
tree-structure encoding $\mb{z}_{k, t}$ is known. In doing so, we
can evaluate the performance of our classification module of the
parser. Note, the POS tag is not provided.

We observe that the F-1 measure of our approach is 2 points worse
than \cite{vinyals2015grammar}; however, when the ground-truth of
$\mb{z}_{k, t}$ is provided, the F-1 measure is 4 points higher
than~\cite{vinyals2015grammar}, which is significant. Therefore,
we attribute the reason for our approach's underperformance to the
fact that our model may not be effective enough to learn to
predict the tree-encoding $\mb{z}_{k, t}$.


\paragraph{Remarks.}
We view the use of unbinding vectors as the main novelty
of our work. In contrast, all other parsers need to input the
words directly. Our \name separates grammar components $\mb{u}_t$
of a sentence from its lexical units $\mb{f}_t$ so that one
author's grammar style can be characterized by unbinding vectors
$\mb{u}_t$ while his word usage pattern can be characterized by
lexical units $\mb{f}_t$. Hence, our parser enjoys the benefit of
aid in learning the writing style of an author since the
regularities embedded in unbinding vectors $\mb{u}_t$ and the
obtained parse trees characterize the writing style of an author.


\section{Conclusion}
\label{sec:Conclusion}

In this paper, we proposed a new ATPL approach for natural
language generation and related tasks. The model has a novel
architecture based on a rationale derived from the use of Tensor
Product Representations for encoding and processing symbolic
structure through neural network computation. In evaluation, we
tested the proposed model on image captioning. Compared to widely
adopted LSTM-based models, the proposed ATPL gives significant
improvements on all major metrics including METEOR, BLEU, and
CIDEr. Moreover, we observe that the unbinding vectors contain
important grammatical information, which allows us to design an
effective POS tagger and constituency parser with unbinding
vectors as input. Our findings in this paper show great promise of
TPRs. In the future, we will explore extending TPR to a variety of
other NLP tasks.


\begin{thebibliography}{99}

\bibitem{tai2015improved}
K.~S. Tai, R.~Socher, and C.~D. Manning, ``Improved semantic
representations
  from tree-structured long short-term memory networks,'' \emph{arXiv preprint
  arXiv:1503.00075}, 2015.

\bibitem{kumar2016ask}
A.~Kumar, O.~Irsoy, P.~Ondruska, M.~Iyyer, J.~Bradbury,
I.~Gulrajani, V.~Zhong,
  R.~Paulus, and R.~Socher, ``Ask me anything: Dynamic memory networks for
  natural language processing,'' in \emph{International Conference on Machine
  Learning}, 2016, pp. 1378--1387.

\bibitem{kong2017dragnn}
L.~Kong, C.~Alberti, D.~Andor, I.~Bogatyy, and D.~Weiss, ``Dragnn:
A
  transition-based framework for dynamically connected neural networks,''
  \emph{arXiv preprint arXiv:1703.04474}, 2017.

\bibitem{smolensky1990tensor}
P.~Smolensky, ``Tensor product variable binding and the
representation of
  symbolic structures in connectionist systems,'' \emph{Artificial
  intelligence}, vol.~46, no. 1-2, pp. 159--216, 1990.

\bibitem{smolensky2006harmonic}
P.~Smolensky and G.~Legendre, \emph{The harmonic mind: From neural
computation
  to optimality-theoretic grammar. Volume 1: Cognitive architecture}.\hskip 1em
  plus 0.5em minus 0.4em\relax MIT Press, 2006.

\bibitem{mao2015deep}
J.~Mao, W.~Xu, Y.~Yang, J.~Wang, Z.~Huang, and A.~Yuille, ``Deep
captioning
  with multimodal recurrent neural networks (m-rnn),'' in \emph{Proceedings of
  International Conference on Learning Representations}, 2015.

\bibitem{vinyals2015show}
O.~Vinyals, A.~Toshev, S.~Bengio, and D.~Erhan, ``Show and tell: A
neural image
  caption generator,'' in \emph{Proceedings of the IEEE Conference on Computer
  Vision and Pattern Recognition}, 2015, pp. 3156--3164.

\bibitem{karpathy2015deep}
A.~Karpathy and L.~Fei-Fei, ``Deep visual-semantic alignments for
generating
  image descriptions,'' in \emph{Proceedings of the IEEE Conference on Computer
  Vision and Pattern Recognition}, 2015, pp. 3128--3137.

\bibitem{andreas2015deep}
J.~Andreas, M.~Rohrbach, T.~Darrell, and D.~Klein, ``Deep
compositional
  question answering with neural module networks,'' \emph{arXiv preprint
  arXiv:1511.02799}, vol.~2, 2015.

\bibitem{yogatama2016learning}
D.~Yogatama, P.~Blunsom, C.~Dyer, E.~Grefenstette, and W.~Ling,
``Learning to
  compose words into sentences with reinforcement learning,'' \emph{arXiv
  preprint arXiv:1611.09100}, 2016.

\bibitem{maillard2017jointly}
J.~Maillard, S.~Clark, and D.~Yogatama, ``Jointly learning
sentence embeddings
  and syntax with unsupervised tree-lstms,'' \emph{arXiv preprint
  arXiv:1705.09189}, 2017.

\bibitem{jurafsky2017Speech}
D.~Jurafsky and J.~H. Martin, \emph{Speech and Language
Processing}, 3rd~ed.,
  2017.

\bibitem{Stanford_Parser_weblink}
C.~Manning, ``Stanford parser,''
  \url{https://nlp.stanford.edu/software/lex-parser.shtml}, 2017.

\bibitem{toutanova2003feature}
K.~Toutanova, D.~Klein, C.~D. Manning, and Y.~Singer,
``Feature-rich
  part-of-speech tagging with a cyclic dependency network,'' in
  \emph{Proceedings of the 2003 Conference of the North American Chapter of the
  Association for Computational Linguistics on Human Language Technology-Volume
  1}.\hskip 1em plus 0.5em minus 0.4em\relax Association for Computational
  Linguistics, 2003, pp. 173--180.

\bibitem{zhu2013fast}
M.~Zhu, Y.~Zhang, W.~Chen, M.~Zhang, and J.~Zhu, ``Fast and
accurate
  shift-reduce constituent parsing.'' in \emph{Proceedings of Annual Meeting of
  the Association for Computational Linguistics (ACL)}, 2013, pp. 434--443.

\bibitem{vinyals2015grammar}
O.~Vinyals, {\L}.~Kaiser, T.~Koo, S.~Petrov, I.~Sutskever, and
G.~Hinton,
  ``Grammar as a foreign language,'' in \emph{Advances in Neural Information
  Processing Systems}, 2015, pp. 2773--2781.

\bibitem{Stanford_Glove_weblink}
J.~Pennington, R.~Socher, and C.~Manning, ``Stanford glove: Global
vectors for
  word representation,'' \url{https://nlp.stanford.edu/projects/glove/}, 2017.

\bibitem{vaswani2017attention}
A.~Vaswani, N.~Shazeer, N.~Parmar, J.~Uszkoreit, L.~Jones, A.~N.
Gomez,
  {\L}.~Kaiser, and I.~Polosukhin, ``Attention is all you need,'' in
  \emph{Advances in Neural Information Processing Systems}, 2017, pp.
  6000--6010.

\bibitem{SCN_CVPR2017}
Z.~Gan, C.~Gan, X.~He, Y.~Pu, K.~Tran, J.~Gao, L.~Carin, and
L.~Deng,
  ``Semantic compositional networks for visual captioning,'' in
  \emph{Proceedings of the IEEE Conference on Computer Vision and Pattern
  Recognition}, 2017.

\bibitem{COCO_weblink}
COCO, ``Coco dataset for image captioning,''
  \url{http://mscoco.org/dataset/#download}, 2017.

\bibitem{he2016deep}
K.~He, X.~Zhang, S.~Ren, and J.~Sun, ``Deep residual learning for
image
  recognition,'' in \emph{Proceedings of the IEEE Conference on Computer Vision
  and Pattern Recognition}, 2016, pp. 770--778.

\bibitem{tensorflow2015-whitepaper}
 M.~Abadi, A.~Agarwal, P.~Barham,
E.~Brevdo, Z.~Chen, C.~Citro, G.~S. Corrado,
  A.~Davis, J.~Dean, M.~Devin, S.~Ghemawat, I.~Goodfellow, A.~Harp, G.~Irving,
  M.~Isard, Y.~Jia, R.~Jozefowicz, L.~Kaiser, M.~Kudlur, J.~Levenberg,
  D.~Man\'{e}, R.~Monga, S.~Moore, D.~Murray, C.~Olah, M.~Schuster, J.~Shlens,
  B.~Steiner, I.~Sutskever, K.~Talwar, P.~Tucker, V.~Vanhoucke, V.~Vasudevan,
  F.~Vi\'{e}gas, O.~Vinyals, P.~Warden, M.~Wattenberg, M.~Wicke, Y.~Yu, and
  X.~Zheng, ``{TensorFlow}: Large-scale machine learning on heterogeneous
  systems,'' 2015, software available from tensorflow.org. [Online]. Available:
  \url{https://www.tensorflow.org/}


\bibitem{papineni2002bleu}
K.~Papineni, S.~Roukos, T.~Ward, and W.-J. Zhu, ``Bleu: a method
for automatic
  evaluation of machine translation,'' in \emph{Proceedings of the 40th annual
  meeting on association for computational linguistics}.\hskip 1em plus 0.5em
  minus 0.4em\relax Association for Computational Linguistics, 2002, pp.
  311--318.

\bibitem{banerjee2005meteor}
S.~Banerjee and A.~Lavie, ``Meteor: An automatic metric for mt
evaluation with
  improved correlation with human judgments,'' in \emph{Proceedings of the ACL
  workshop on intrinsic and extrinsic evaluation measures for machine
  translation and/or summarization}.\hskip 1em plus 0.5em minus 0.4em\relax
  Association for Computational Linguistics, 2005, pp. 65--72.

\bibitem{vedantam2015cider}
R.~Vedantam, C.~Lawrence~Zitnick, and D.~Parikh, ``Cider:
Consensus-based image
  description evaluation,'' in \emph{Proceedings of the IEEE Conference on
  Computer Vision and Pattern Recognition}, 2015, pp. 4566--4575.

\bibitem{Penn_treebank_weblink}
M.~P. Marcus, B.~Santorini, M.~A. Marcinkiewicz, and A.~Taylor,
``Penn
  treebank,'' \url{https://catalog.ldc.upenn.edu/ldc99t42}, 2017.

\end{thebibliography}

\end{document}